\documentclass{article} 
\usepackage{iclr2025_conference,times}
\usepackage{bandai}

\usepackage{hyperref}
\usepackage{url}
\usepackage{colortbl}
\usepackage{enumitem}
\usepackage{framed}
\usepackage{tcolorbox}
\usepackage{setspace}
\usepackage{array}
\usepackage{caption}
\usepackage{subcaption}
\usepackage{multirow}  
\usepackage{booktabs}

\usepackage{tabularx}
\usepackage{xcolor}
\tcbset{
  colback=blue!5!white, colframe=blue!40!black,
  boxrule=0.3pt, arc=2pt, left=3pt, right=3pt, top=1.5pt, bottom=1.5pt
}
\newcommand{\Tag}[1]{\textsf{\bfseries #1}}

\usepackage{xcolor}

\usepackage{fancyhdr}
\usepackage{ifthen}

\fancypagestyle{firstpagestyle}{
  \fancyhf{}
  \fancyhead[L]{\includegraphics[width=70pt]{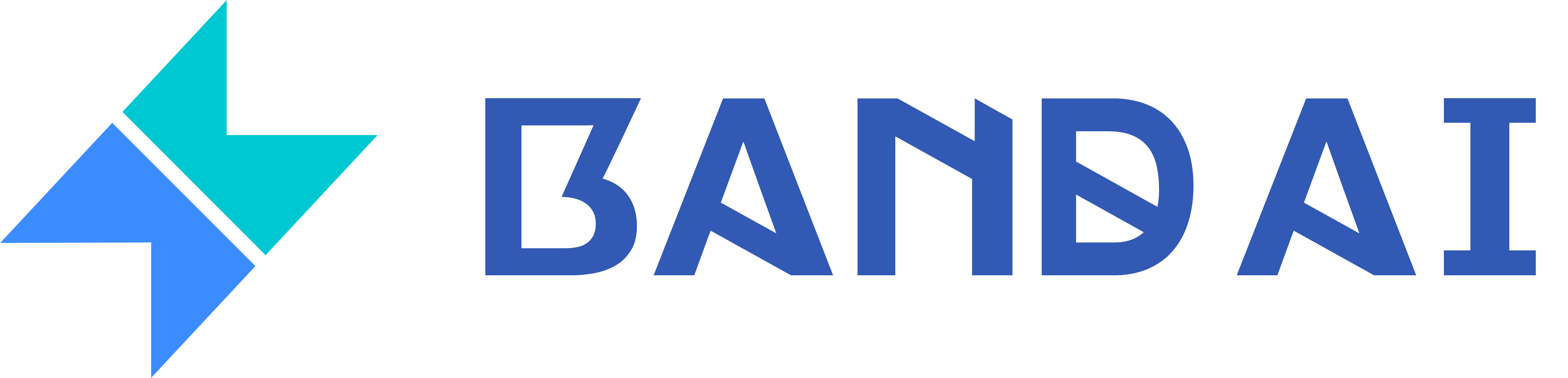}}
}

\definecolor{HeadingColor}{HTML}{325AB4}
\definecolor{MyLightBlue}{HTML}{78E6DC}

\title{\titlefont ReportBench: Evaluating Deep Research Agents  via Academic Survey Tasks}

\author{Minghao Li\thanks{Equal contribution}\ \thanks{Corresponding author},\, Ying Zeng\footnotemark[1],\, Zhihao Cheng\footnotemark[1],\, Cong Ma,\, Kai Jia \\
ByteDance BandAI\symbolbandai\\
\small{\texttt{\{liminghao.bd,zengying.ss,zhihao.cheng,macong.13,jiakai\}@bytedance.com} }\\
}

\iclrfinalcopy 
\begin{document}

\maketitle

\begin{abstract}
The advent of Deep Research agents has substantially reduced the time required for conducting extensive research tasks. However, these tasks inherently demand rigorous standards of factual accuracy and comprehensiveness, necessitating thorough evaluation before widespread adoption. 
In this paper, we propose ReportBench, a systematic benchmark designed to evaluate the content quality of research reports generated by large language models (LLMs). 
Our evaluation focuses on two critical dimensions: (1) the quality and relevance of cited literature, and (2) the faithfulness and veracity of the statements within the generated reports. 
ReportBench leverages high-quality published survey papers available on arXiv as gold-standard references, from which we apply reverse prompt engineering to derive domain-specific prompts and establish a comprehensive evaluation corpus. 
Furthermore, we develop an agent-based automated framework within ReportBench that systematically analyzes generated reports by extracting citations and statements, checking the faithfulness of cited content against original sources, and validating non-cited claims using web-based resources. 
Empirical evaluations demonstrate that commercial Deep Research agents such as those developed by OpenAI and Google consistently generate more comprehensive and reliable reports than standalone LLMs augmented with search or browsing tools. However, there remains substantial room for improvement in terms of the breadth and depth of research coverage, as well as factual consistency. 
The complete code and data will be released at the following link: \url{https://github.com/ByteDance-BandAI/ReportBench}.
\end{abstract}

\section{Introduction}
The rapid development of LLM-powered Deep Research agents has revolutionized the process of knowledge synthesis by enabling autonomous execution of extensive research tasks, including academic literature surveys, industry analyses, and market assessments~\cite{chen2025ai4research, DBLP:journals/corr/abs-2502-18864, DBLP:journals/corr/abs-2408-06292, DBLP:journals/corr/abs-2505-18705, DBLP:journals/corr/abs-2504-08066, DBLP:journals/corr/abs-2504-03160, DBLP:journals/corr/abs-2504-21776}. Tasks that traditionally required days or weeks of manual effort can now be completed within minutes. Notable examples include advanced systems such as OpenAI~\cite{openai_deep_research} and Google’s Gemini Deep Research~\cite{google_deep_research}, which effectively integrate various external tools and perform multiple rounds of deep reasoning. Despite their promising capabilities, widespread practical adoption critically depends on their ability to consistently deliver research reports with high factual accuracy and comprehensive content quality. Therefore, it is essential to monitor and ensure the quality of generated reports through evaluation. However, defining what constitutes a good report is challenging and lacks broad consensus, resulting in the current absence of mature evaluation methodologies for research report generation.

In addressing this challenge, we decompose the evaluation of research reports generated by LLMs into two core dimensions: writing quality and report content. Due to the subjectivity of writing-style evaluation, while the criteria for assessing content quality can be more clearly defined, this work focuses primarily on the evaluation of report content, leaving the assessment of writing quality to future work. Specifically, we assert that the content quality of research reports hinges on two critical factors: (1) the quality and relevance of cited literature, and (2) the faithfulness and veracity of generated statements, whether derived from cited references or produced by the model.

To establish a high-quality benchmark capable of rigorously assessing research reports, we propose ReportBench, a novel evaluation framework leveraging expert-generated literature reviews. Given the constraints of relying on human annotators, who typically vary in expertise, we propose using published survey papers available on \href{https://arxiv.org/}{arXiv} as gold-standard references. Published survey papers are typically written by domain experts and have undergone a peer review process that provides additional expert-level validation, considered among the highest-quality research reports currently available.

In practice, our methodology unfolds in two phases. First, we generate domain-specific retrieval prompts directly from expert-authored survey papers on arXiv: by analyzing each paper’s publication date and full text, we generate three granularity levels of prompts (sentence-level, paragraph-level, and richly detailed versions) that precisely capture the scope, methods, and temporal constraints of the original research. These prompts form the backbone of our evaluation corpus, ensuring that downstream agents search and synthesize information within the exact topical and chronological boundaries of each survey. We extract the list of cited references from the arXiv surveys as the ground truth. Given the synthesized prompts as test inputs, Deep Research agents conduct research and generate reports, which are then evaluated based on the reference overlap with the ground truth, serving as a measure of the research skills.

In the second phase of our validation pipeline, we design two different verification procedures based on whether a statement includes an explicit citation to external literature. Specifically, for cited statements, the system identifies all in-text citations within the report, maps each citation to its corresponding source document, and employs semantic matching to ensure factual support from the cited literature. For non-cited statements, the framework employs a voting mechanism across multiple web-connected models to verify the factuality of these statements. By combining these complementary validation procedures, ReportBench delivers a systematic and detailed assessment of AI-generated research reports, ensuring the relevance and quality of cited literature and the factual accuracy of all claims through citation-based and web-based validation.

Our contributions can be summarized as follows:
\begin{itemize}[leftmargin=2em]
\item We present \textbf{ReportBench}, a systematic benchmark designed to evaluate the quality of research reports generated by Deep Research agents, with a focus on the quality of references and the factual accuracy of all statements presented in the report.
\item We propose an automated and scalable data synthesis method for constructing academic survey tasks, including prompts and ground truth, from expert-authored survey papers on arXiv. Besides, we introduce an automatic agentic evaluation framework that evaluates the precision and recall of the generated report with respect to a ground-truth reference and performs factual verification of individual claims made within the report.
\item We release a comprehensive benchmark suite—datasets, prompts, and evaluation scripts—to support reproducible research and community-driven progress in evaluating LLM-based knowledge synthesis.
\end{itemize}

\begin{figure}[!htbp]
  \centering
  \includegraphics[width=\textwidth]{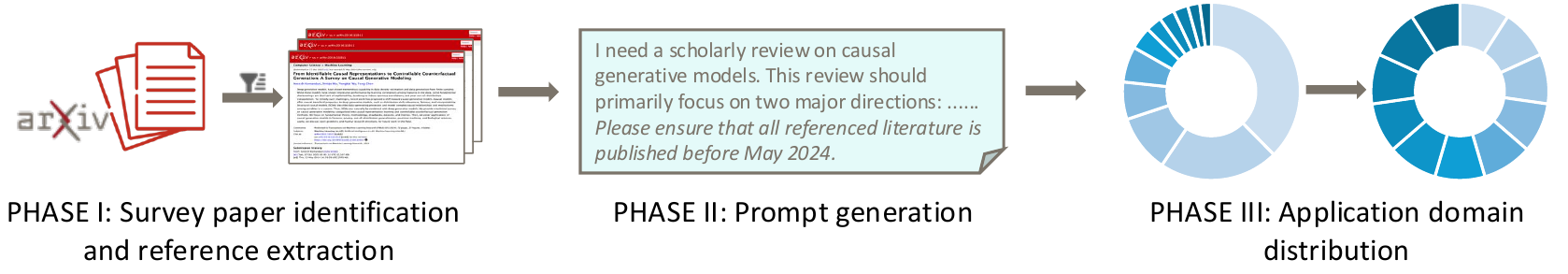}
  \caption{Overall benchmark data construction workflow.}
  \label{fig:data_process_workflow}
\end{figure}

\section{Methodology}
We introduce \textbf{ReportBench}, a comprehensive evaluation framework designed to rigorously assess Deep Research agents through two interconnected components: (i) the automated construction of high-quality benchmark datasets derived from expert-authored survey papers, and (ii) a systematic validation pipeline that evaluates the quality and factual consistency of AI-generated research reports. In the following sections, we detail the processes that underlie the synthesis of the dataset and the design of our evaluation workflow.

\subsection{Dataset Construction}
\label{subsec:dataset_construction}
In this section, we detail the end-to-end pipeline to construct high–quality deep research questions along with ground-truth answers based on published survey papers. This workflow comprises three consecutive phases: (i) survey paper identification and reference extraction, (ii)~prompt generation, and (iii)~application domain distribution. A diagram illustrating the data construction process is presented in Figure~\ref{fig:data_process_workflow}.

\subsubsection{Phase~I: Survey Paper Identification and Reference Extraction}
\label{subsubsec:survey_identification}
The first step is to identify high-quality survey papers to create problems. We start from the complete \href{https://arxiv.org/}{arXiv} metadata snapshot~\cite{arxiv_org_submitters_2024} and only reserve papers with a submission date later than \texttt{2020-01-01}. To ensure the quality of papers, we only select those that have undergone peer review and have been formally published. We achieve this by using regular expressions, \emph{i.e.}, querying over titles to match \emph{``survey''} or \emph{``review''} to filter survey papers and searching \emph{``published''} or \emph{``accepted''} in the comments field of a submission. To reduce systematic false positives in domains such as astronomy, we prompted \texttt{GPT-4o}~\cite{hurst2024gpt} with each paper’s title and abstract to produce a binary classification of whether the paper is a literature survey.

For each survey paper, we analyze its LaTeX source file to extract cited references. Specifically, we parse LaTeX citation commands, identify and retrieve relevant bibliographic entries from associated bibliography databases, and filter these to retain only references explicitly cited in the main text. Hence, the extracted bibliography mirrors the true citation pattern of the paper. The resulting dataset constitutes a gold-standard benchmark for evaluating retrieval precision. Finally, we ultimately retained 678 papers.

\subsubsection{Phase II: Prompt Generation}
\label{subsubsec:prompt_generation}
Survey papers can be regarded as a great depth of research work focused on a specific topic at a specific time, making it possible to create deep research questions via a \emph{reverse prompt engineering} manner. In other words, given the publication date and the full text of a survey paper obtained through a PDF parsing tool, we prompt an LLM to generate a query whose ideal answer is precisely that paper. Hence, we obtain a query and its ground truth (the survey paper itself). To increase the diversity of prompts, we design three types of prompt templates:
\begin{tcolorbox}
[notitle,boxrule=0pt,colback=MyLightBlue!10,colframe=MyLightBlue!10]
\texttt{\fcolorbox{white}{HeadingColor!10}{\textbf{Sentence-level prompt}} \\
A single sentence that succinctly defines the overarching academic field covered by the survey.\\
\fcolorbox{white}{HeadingColor!10}{\textbf{Paragraph-level prompt}}\\
A short paragraph elaborating the research area, its main subtopics, and the methodological perspectives covered in the survey.\\
\fcolorbox{white}{HeadingColor!10}{\textbf{Detail-rich prompt}}\\
A detailed question that comprehensively describes the specific research domain, key research directions, and the methodological approaches of interest. Additional constraints may be included, such as preferred conferences or journals, language of the cited literature (e.g., English, Chinese), participating institutions or laboratories.}
\end{tcolorbox}
Besides, to ensure temporal consistency, we require each generated prompt to include a cut-off date corresponding to the most recent update of the paper. For example, an expression like the following is needed.
\begin{quote}
  ``Ensure only papers published before April 2025 are referenced.'' 
\end{quote}
This requirement ensures that LLMs’ retrieval window matches the survey’s citation horizon and prevents leakage of post-publication knowledge and hacking of finding the exact same paper. Nevertheless, we still observe a phenomenon akin to prompt hacking during model evaluation, \emph{i.e.}, the model disregards the imposed temporal constraints and directly retrieves the original source paper. To address this issue, we augment the prompt with an additional explicit instruction, stipulating that the model must refrain from citing the original paper corresponding to the given prompt. We present three prompt examples in Appendix~\ref{sec:app:promp_exmp}

\subsubsection{Phase III: Application Domain Distribution}
\label{subsubsec:application_domain_classification}
To facilitate a more granular analysis of tested models, we classified the prompts into distinct application domains. Specifically, we utilize Gemini 2.5 Pro \cite{comanici2025gemini} to classify each paper based on the title and abstract. This process yields ten distinct categories, as shown in the following box. To reduce misclassification, we introduce an unknown category, allowing the model to assign uncertain cases to this class.
{\footnotesize
\begin{tcolorbox}[colback=black!2, colframe=black!15, boxrule=0.2pt, enlarge left by=0mm]
\begin{tabularx}{\linewidth}{@{} l X @{\hspace{1em}} l X @{}}
  \Tag{A} & Basic Research and Scientific Exploration &
  \Tag{F} & Transportation and Smart Mobility \\
  \Tag{B} & Information and Communications Technology &
  \Tag{G} & Public Safety and Social Governance \\
  \Tag{C} & Artificial Intelligence and Data Intelligence &
  \Tag{H} & Finance and Business Services \\
  \Tag{D} & Healthcare and Biomedicine &
  \Tag{I} & Energy and Environmental Sustainability \\
  \Tag{E} & Manufacturing and Smart Manufacturing &
  \Tag{J} & Culture, Media, and Digital Content \\
  \Tag{K} & Unknown Category &
          &
\end{tabularx}
\end{tcolorbox}
}

\begin{figure}[!htbp]
  \centering
  \begin{subfigure}{0.46\linewidth}      
    \centering
    \includegraphics[width=\linewidth]{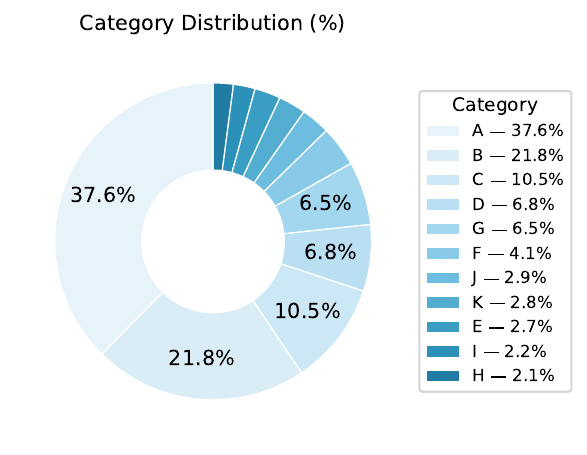}
    \caption{Category distribution (pie).}
    \label{fig:pie}
  \end{subfigure}
  \hfill                                    
  \begin{subfigure}{0.46\linewidth}
    \centering
    \includegraphics[width=\linewidth]{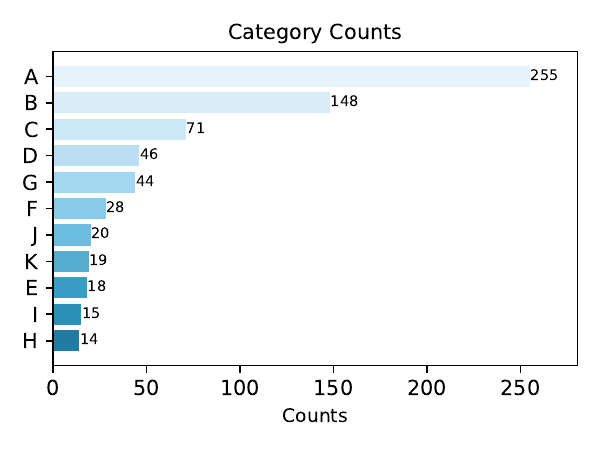}
    \caption{Category counts (bar).}
    \label{fig:bar}
  \end{subfigure}

  \caption{Application domain distribution of the 678 filtered ReportBench prompts: (a) a pie chart showing the proportion of each application domain, (b) a bar chart illustrating the total task counts across all 11 categories.}
  \label{fig:pie_bar_combo}
\end{figure}

The distribution of prompts across these domains is inherently biased due to the specific disciplinary focus of the arXiv corpus, as shown in Figure~\ref{fig:pie_bar_combo}. To create a balanced and general test set, we down-sample a total of 100 papers. As we have mentioned before, we create three types of prompts for each paper. Thus, we randomly sample from these three types to obtain the final prompt with diversity. In other words, a dataset with 100 prompts is created, which we name as \textbf{ReportBench}. The quality of the classification of this subset was then reviewed and validated by four research experts.

\subsection{Evaluation Process}
\begin{figure}[!htbp]
  \centering
  \includegraphics[width=\textwidth]{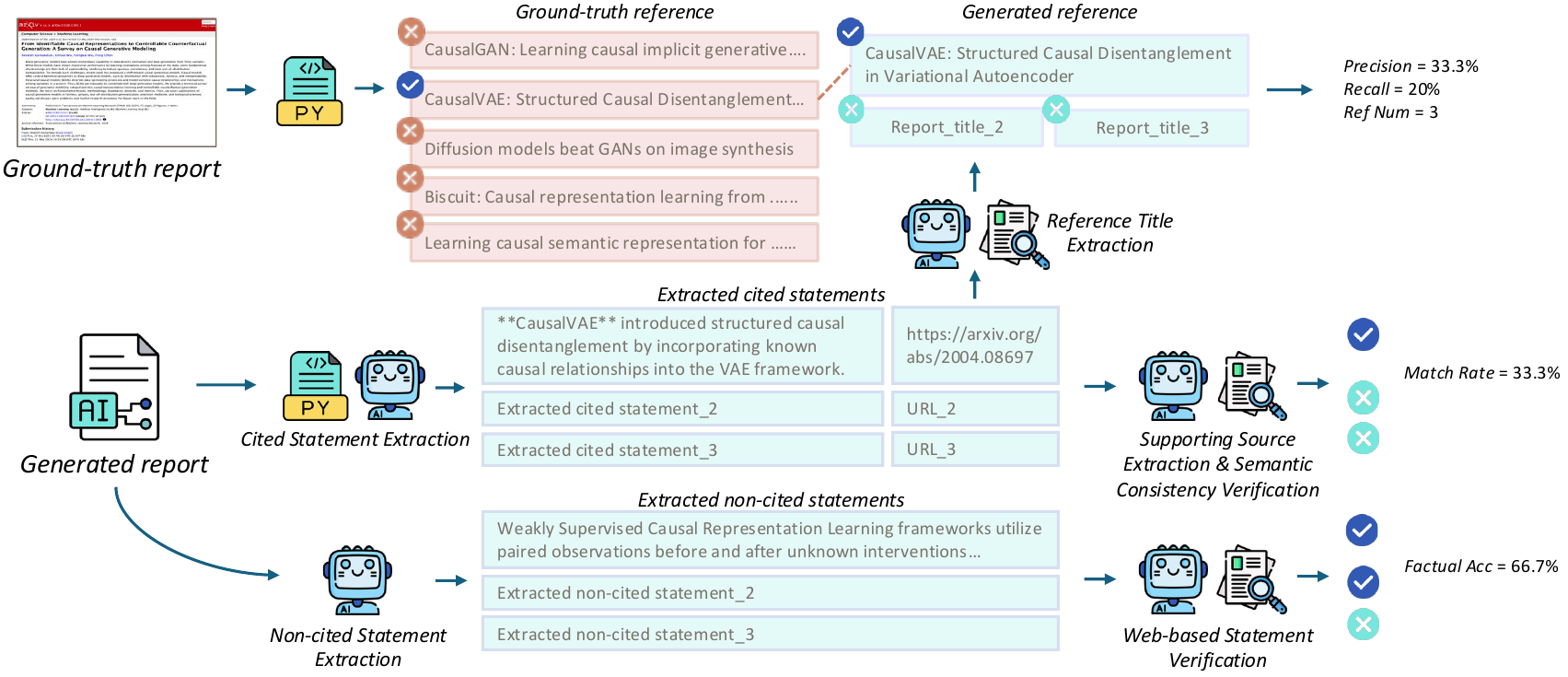}
  \caption{Evaluation Process.}
  \label{fig:eval_workflow}
\end{figure}

Our evaluation process uses test prompts derived from reverse prompt engineering, which require models to generate complete research reports under two constraints: a time limit and a restriction against referencing the original report, which is presented in Figure~\ref{fig:eval_workflow}. \textbf{Content quality} is first evaluated by assessing the cited references: we compare the reference list in the generated report with that of the ground truth, and the overlap ratio between the two lists serves as an indicator of the report’s overall quality. \textbf{Statement factuality} is further assessed through two complementary validation procedures. For cited statements, we verify alignment with source documents via semantic matching, while for non-cited statements, we adopt a multi-model voting mechanism to assess factual correctness. This dual strategy ensures both the faithfulness of cited content and the veracity of non-cited claims in evaluating Deep Research reports.

\paragraph{Content quality}
We first extract all URLs from the report. Since most reports generated by the Deep Research products use URL links to cite web pages, we adopt the same citation format throughout our evaluation, including when assessing the base model. While this approach results in longer text, it offers the advantage of placing the citation immediately adjacent to the corresponding statement, which ensures consistent performance even under chunked evaluation settings. After normalizing and deduplicating them, we retrieve the content of each web page. An LLM is then used to determine whether each page corresponds to a scholarly article and, if so, to extract the article title. Finally, we compute the overlap between the extracted document titles and the ground-truth reference titles to produce a quality score.

\paragraph{Cited statements}
We design a three-stage structured validation pipeline. First, an LLM automatically identifies all statements in the generated report that contain explicit citation links, establishing a mapping between each statement and its referenced source. Second, we retrieve the full content of each cited webpage via web scraping and prompt the LLM to locate the most semantically relevant passage that supports the original statement. Finally, the LLM performs consistency verification by comparing the statement with the retrieved content, and the results are aggregated to compute an overall citation consistency score for the report. Unlike traditional “LLM-as-a-judge” approaches, which often suffer from instruction-following issues or biased scoring, our method decomposes the evaluation into fine-grained, interpretable, and verifiable steps. All intermediate outputs are retained for optional human inspection, thereby maximizing the reliability and transparency of the evaluation process.

\paragraph{Non-cited statements}
We use a simple two-step validation process. First, we extract all factual statements in the report that do not have any citations, and remove content that is general common sense or already supported by references. Then, we ask several web-connected LLMs to check each statement independently. Each model looks up information online and gives its judgment. We combine their answers using a voting mechanism to decide whether the statement is likely to be correct. This approach avoids relying on a single model and makes the validation more reliable.

\section{Experiment}
In this section, we present the performance of a diverse set of models evaluated on ReportBench. Specifically, we examine specialized Deep Research agents from OpenAI and Google Gemini. Additionally, we assess several state-of-the-art (SOTA) base models, originally lacking native Internet access, by augmenting them with an external search engine and link reader to enable the web-retrieval capabilities essential for completing our evaluation tasks. These enhanced base models are then benchmarked alongside the native Deep Research agents.

\subsection{Setttings}
In our evaluation workflow, different LLMs are used for different components. For statement extraction, supporting source extraction, and semantic consistency verification, we adopt \texttt{gpt-4o}. For the fact-checking of non-cited statements, we employ two web-connected models: \texttt{gemini-2.5-pro} and \texttt{gemini-2.5-flash}. Each model performs three independent judgments per statement, resulting in a total of six verdicts. The final decision is determined by majority voting, and the proportion of votes is recorded as a confidence score. In the evaluation of base models, we integrated search and link-reading tools using each model’s native function call interface. Specifically, we used SerpAPI\footnote{\url{https://serpapi.com/}} for Google Search access and Firecrawl\footnote{\url{https://www.firecrawl.dev/}} for retrieving web pages in Markdown format. Due to context length limitations, we capped the maximum number of tool calls at five per instance.

To evaluate the performance of both Deep Research agents and base models, we manually collected responses from the web-based interfaces of OpenAI and Gemini, as well as batch-executed outputs from the base models, during the period from July 14 to July 25. During data collection, we ensured that OpenAI was using the standard version of Deep Research, powered by the \texttt{o3} model. For Gemini, we made sure that both the “Gemini 2.5 Pro” and “Deep Research” toggles were enabled on the web interface to activate its full research capabilities.

\subsection{Evaluation Metrics}  

As described in our evaluation logic, we define three sets of metrics to assess a model’s performance in conducting scientific research tasks. First, we compute the \textbf{precision} and \textbf{recall} of retrieved references against groundtruth references. Precision reflects the proportion of cited references that are relevant, while recall measures the proportion of ground-truth references successfully retrieved. We also report the average number of references per report to capture the model’s reference density. To evaluate statement-level performance, we measure the average number of cited statements and non-cited statements per report. For cited statements, we compute the \textbf{match rate}, i.e., the proportion of statements that are semantically consistent with their cited sources. For non-cited statements, we compute the \textbf{factual accuracy}, defined as the proportion of statements that are verified to be factually correct via web-connected LLMs.




\setlength{\tabcolsep}{3pt}         
\renewcommand{\arraystretch}{1.5}   

\begin{table}[htbp]
\footnotesize
\centering
\begin{tabular}{@{}c|ccc|cc|cc@{}}
\toprule
\multirow{2}{*}{\textbf{Test Model}} & \multicolumn{3}{c|}{\textbf{Reference}} & \multicolumn{2}{c|}{\textbf{Cited statements}} & \multicolumn{2}{c}{\textbf{Non-cited statements}} \\
 
\cline{2-8}
 & Precision & Recall & Ref Num & Match Rate & Count & Factual Acc & Count \\
\hline
OpenAI Deep Research   & \textbf{0.385} & 0.033 & 9.89  & \textbf{78.87\%} & 88.2  & 95.83\% & 38.9  \\
Gemini Deep Research   & 0.145          & \textbf{0.036} & 32.42 & 72.94\% & 96.2  & 92.21\% & 49.6  \\
\hline
\hline
gemini-2.5-flash       & 0.237 & 0.012 & 5.47  & 44.88\% & 12.1  & \textbf{98.52\%} & 11.5  \\
gemini-2.5-pro         & 0.269 & 0.010 & 4.27  & 59.24\% & 6.58  & 96.08\% & 9.35  \\
openai-o3              & 0.299 & 0.031 & 12.26 & 31.43\% & 16.16 & 82.22\% & 11.51 \\
claude4-sonnet         & 0.337 & 0.021 & 6.74  & 73.67\% & 14.93 & 92.64\% & 17.07 \\
\bottomrule
\end{tabular}
\caption{Performance metrics of OpenAI Deep Research, Gemini Deep Research, and the base models. ``Ref Num'' denotes the average number of references per report, and ``Count'' denotes the average number of cited or non-cited statements.}
\label{tab:openai_deep_research_metrics}
\end{table}

\subsection{Product-Level Comparative Analysis}

Table~\ref{tab:openai_deep_research_metrics} presents the performance metrics of OpenAI Deep Research and Gemini Deep Research. In terms of retrieval performance, OpenAI achieves significantly higher precision (0.385) compared to Gemini (0.145), indicating that the references it retrieves are more likely to match the gold-standard set. Gemini shows a slightly higher recall (0.036 vs. 0.033), but this gap is negligible in practical terms. As shown in the table, Gemini generates more than three times as many cited statements on average (32.42 vs. 9.89), yet this increase does not translate into a significant improvement in recall. This suggests that Gemini tends to over-generate citations without proportionally improving the coverage of high-quality references. In some cases, excessive citation may even introduce redundancy or dilute the relevance of retrieved content. Given that the ground truth from ReportBench includes an average of 153 references per paper, with many citations supporting the same or overlapping statements, we believe recall should be considered a secondary signal rather than the primary focus of evaluation.

In terms of statement quality, both products demonstrate strong performance on generating reports, with OpenAI achieving better citation alignment (Match Rate 78.87\% vs. 72.94\%). Despite citing fewer sources, OpenAI maintains a high average alignment score (88.2), suggesting stronger precision in citation usage. For non-cited statements, Gemini produces more such content (49.6 vs. 38.9), while OpenAI achieves better factual accuracy (95.83\% vs. 92.21\%), indicating its stronger calibration in generating reliable citation-free content.

\subsection{Model-Level Comparative Analysis}

We now analyze the results across several foundation models and compare them with the corresponding Deep Research agents.

\textbf{OpenAI Deep Research vs. \texttt{o3}} \\
OpenAI Deep Research and \texttt{o3} exhibit similar retrieval performance, with precision (0.385 vs. 0.299) and recall (0.033 vs. 0.031) showing only slight differences. Meanwhile, the average number of references per report is also comparable (9.89 vs. 12.26). This observation aligns well with OpenAI’s official disclosure that the retrieval and synthesis backbone of Deep Research is powered by the \texttt{o3} model.

However, we observe substantial differences in the number and quality of generated statements. OpenAI Deep Research produces significantly more cited statements on average (88.2 vs. 16.16) and more non-cited statements (38.9 vs. 11.51), while achieving a notably higher citation match rate (78.87\% vs. 31.43\%) and factual accuracy (95.83\% vs. 82.22\%). This suggests that Deep Research is not a direct output of \texttt{o3}, but rather likely incorporates an additional writing module, possibly optimized via fine-tuning or structured pipelines. Such a pipeline may be responsible for structuring retrieved content into a more coherent, citation-aligned report.

\textbf{Gemini Deep Research vs. \texttt{gemini-2.5-pro}} \\
Similarly, Gemini Deep Research and its base model \texttt{gemini-2.5-pro} diverge significantly across multiple dimensions. Gemini Deep Research trades off some precision (0.145 vs. 0.269) to achieve much higher recall (0.036 vs. 0.010) and generates far more references per report (32.42 vs. 4.27). In terms of statement volume, it produces many more cited statements (96.2 vs. 6.58) and non-cited statements (49.6 vs. 9.35). Despite this increase in volume, its citation alignment remains strong (72.94\% vs. 59.24\%), while its non-cited statement accuracy is slightly lower than the base model (92.21\% vs. 96.08\%). These pronounced gaps—in precision/recall trade-off, citation count, and overall coverage—mirror the contrast observed between OpenAI Deep Research and \texttt{o3}, and suggest that the system has undergone targeted optimization for thorough research and report generation. Taken together with the visible “plan” and “step-by-step reasoning” phases presented in the Gemini Deep Research web interface, it seems plausible that the system functions more like a thoughtfully constructed multi-agent workflow or pipeline.

\textbf{Base-Model Comparison} \\
Among the four base models, \texttt{claude4-sonnet} demonstrates the most balanced performance—achieving a precision of 0.337, a recall of 0.021, an average of 6.74 reference documents per report, a high citation semantic consistency (73.67\%), and a strong non-cited statement factual accuracy (92.64\%). In contrast, \texttt{gemini-2.5-pro} attains higher precision (0.269) at the expense of recall (0.010) and generates fewer reference documents on average (4.27 per report), limiting its coverage. \texttt{gemini-2.5-flash} underperforms on both precision (0.237) and recall (0.012), with lower citation semantic consistency (44.88\%), indicating poorer citation relevance. Meanwhile, \texttt{o3} produces the most references (12.26 per report) and moderate recall (0.031), but its citation semantic consistency (31.43\%) and non-cited statement accuracy (82.22\%) lag behind. 

Overall, Deep Research products significantly outperform their base models in coverage and factual grounding, pointing to the value of task-specific model fine-tuning or pipeline design beyond standalone LLM capabilities. 

\section{Analysis}
It is notable that many models exhibit low citation semantic consistency, particularly when relying on function-call mechanisms to retrieve and cite literature. In our manual inspection of evaluation results, we identified two representative failure types: \textbf{statement hallucination}, where the content deviates from the cited source, and \textbf{citation hallucination}, where the reference itself is fabricated.

\paragraph{Statement Hallucination.}  In our manual audit of arXiv:2407.15186 test cases, we identified representative errors in statement generation. For example, OpenAI Deep Research generated the following claim:

\begin{quote}
Kulkarni \textit{et al.} (2025) and others introduced RL fine‐tuning where the model gets a reward of +1 if its SQL yields the correct answer when run, and 0 otherwise (\href{https://arxiv.org/html/2503.23157v2#section.3.2}{arXiv:2503.23157v2, §3.2}).
\end{quote}

Upon inspection, the cited part indeed describes a reasoning‐enhanced RL reward scheme for Text‐to‐SQL; however, the list of authors does not include “Kulkarni.”. In fact, Kulkarni did publish a paper on reinforcement learning and Text-to-SQL, but it was not among the references cited in the generated report. We speculate that the model may have encountered similar data during training and mistakenly attributed Kulkarni’s contribution to this cited paper.

\paragraph{Citation Hallucination.} During our evaluation of arXiv:2009.12619, we observed a clear instance of link hallucination in the generated report from \texttt{gemini-2.5-pro}. The model generated the claim:

\begin{quote}
In-vehicle Crowd Monitoring: The use of surveillance cameras inside buses and trains for passenger counting is a well-established practice. Advanced image processing and computer vision techniques can automatically analyze video feeds to estimate the passenger load. For instance, a system was proposed to estimate the number of passengers in a bus using image processing techniques on the captured video frames, achieving high accuracy. 
[Vision-Based In-Vehicle Crowd Monitoring](\url{https://www.researchgate.net/publication/224217198_A_vision-based_system_for_in-vehicle_crowd_monitoring}).
\end{quote}

However, the cited URL does not exist and appears to be entirely fabricated by the model. Because the link cannot be resolved, no supporting text or evidence can be retrieved to validate the statement, resulting in a citation mismatch. This example highlights a common error mode in function-call–driven retrieval: the model confidently invents plausible-looking reference links that nonetheless point to nothing, undermining factual grounding. 

These examples demonstrate that even advanced Deep Research agents remain susceptible to hallucinating author names, misaligning citations, and fabricating links. Crucially, our evaluation metrics—especially citation semantic consistency—are sensitive to such discrepancies, allowing us to quantitatively capture and penalize these hallucination phenomena across model outputs.

\section{Related Work}

Long-standing interest has been in the use of AI to synthesize information, not only in the writing of scientific articles~\cite{chen2025ai4research, DBLP:journals/corr/abs-2502-18864, DBLP:journals/corr/abs-2408-06292, DBLP:journals/corr/abs-2505-18705, DBLP:journals/corr/abs-2504-08066}, but also in the search for information and the generation of reports in the general domains~\cite{DBLP:journals/corr/abs-2504-03160, DBLP:journals/corr/abs-2504-21776}. With the rapid advancement of information synthesis research, the evaluation of long-form reports has become increasingly important. Existing evaluation benchmarks focus primarily on individual aspects such as fact checking, citation judgment, or overall writing quality. Although these benchmarks can assess the quality of the report to some extent, several limitations remain unaddressed.

\textbf{Fact Checking Evaluation} Driven by efforts from both academia and industry, automated fact checking has evolved into a well-established multistage pipeline, which has become the dominant research paradigm in the field~\cite{eldifrawi-etal-2024-automated}. Claim detection aims to identify factual statements worth verifying from large volumes of text~\cite{DBLP:journals/tacl/GuoSV22, DBLP:journals/nlpj/PanchendrarajanZ24}, while evidence retrieval focuses on retrieving relevant documents or textual snippets that support or refute a given claim~\cite{eldifrawi-etal-2024-automated, DBLP:conf/ecir/NanhekhanVMVSA25}. Building on this pipeline, several benchmarks have been proposed to evaluate the performance of fact checking in both the general domain~\cite{thorne-etal-2018-fever, DBLP:conf/acl/Ma0WCWLW24} and the scientific domain~\cite{DBLP:conf/emnlp/WaddenLLWZCH20, wadden-etal-2022-scifact, DBLP:journals/corr/abs-2506-10486}. However, these benchmarks focus solely on fact-checking components, rather than evaluating the synthesized information as a whole, limiting their ability to assess recent long-form output from large language models (LLMs), such as full research reports.

\textbf{Citation Evaluation} Research reports often include a substantial amount of citation-related content, and evaluating the precision and standardization of these citations plays a crucial role in assessing the overall quality of the report~\cite{sarol2024assessing}. Given a report with citation content, tasks such as cited context identification, evidence sentence retrieval, and citation accuracy classification are commonly used to analyze citation quality~\cite{sarol2024assessing}. Widely applied in assisted paper writing and review systems, citation verification tools are designed from multiple perspectives, including syntactic verification, existence verification, and semantic verification~\cite{barrot2025trinka, bairagi2024revolutionizing}. While citation correctness and existence have been well-studied, the aspect of citation completeness—i.e., whether all relevant prior work on a given research topic has been cited—remains underexplored. To address this gap, our ReportBench incorporates large language models and search tools to evaluate and verify the completeness of related work coverage, offering a more comprehensive perspective for research report quality assessment.

\textbf{Survey Generation} With the advent of LLMs, automated survey generation has seen rapid progress. Early works leveraged LLMs to improve literature comprehension and survey writing \cite{wang2024autosurveylargelanguagemodels, hu2025taxonomytreegenerationcitation}, achieving better coherence compared to sentence extraction methods. Subsequent research explored structured and hierarchical organization, such as hierarchical catalogue generation with semantic and structural metrics, though these remained limited to outline generation with fixed references. Other approaches focused on modeling paper relationships via citation networks, including AutoSurvey \cite{wang2024autosurveylargelanguagemodels} with a two-stage LLM pipeline and HiReview \cite{hu2025taxonomytreegenerationcitation} with a taxonomy-driven framework, though both faced limitations in capturing human writing styles or relying on restricted citation scopes. More recently, SurveyForge \cite{yan2025surveyforgeoutlineheuristicsmemorydriven} combines human outline structure analysis with high-quality literature retrieval, generating and refining full survey content through a scholar navigation agent. The accompanying SurveyBench evaluates generated surveys across reference, outline, and content quality, showing significant improvements over prior methods.
Compared with SurveyBench, ReportBench focuses solely on well-defined and automatically verifiable dimensions of evaluation—namely, factual faithfulness and correctness. In addition, through an automated construction pipeline, it ensures data quality while offering clear scalability advantages, enabling it to serve as a potential source of training data for targeted report optimization in future work.

\textbf{Deep Research Evaluation} The rise of deep research agents (DRAs), driven by powerful models such as ChatGPT~\cite{openai_deep_research} and Gemini~\cite{google_deep_research}, has underscored the urgent need for robust and targeted evaluation methodologies. While existing benchmarks evaluate capabilities such as web retrieval~\cite{DBLP:journals/corr/abs-2504-12516, DBLP:journals/corr/abs-2504-19314, DBLP:conf/acl/0007Y0WXFZ0ZXH25}, multi-hop factual reasoning~\cite{DBLP:journals/corr/abs-2411-04368, DBLP:conf/iclr/MialonF0LS24, DBLP:journals/corr/abs-2501-14249}, and end-to-end report generation~\cite{DBLP:journals/corr/abs-2506-11763, DBLP:journals/corr/abs-2506-06287}, they often operate at a surface level and fall short of evaluating the core competencies essential for rigorous and reliable research. In contrast, our ReportBench is specifically designed to assess two critical pillars of trustworthy and practical DRA outputs: factual accuracy and citation behavior.

\section{Conclusion}

In this paper, we present \textbf{ReportBench}, a comprehensive benchmark for evaluating the quality of references and the factual accuracy of all statements in reports generated by Deep Research agents. By leveraging expert-authored survey papers as ground-truth and reverse prompt engineering, we enable consistent evaluation of AI-generated research reports across multiple dimensions. Our framework introduces a fine-grained validation workflow that separately assesses cited and non-cited statements, combining citation semantic consistency checks and web-based factual verification. Through large-scale experiments on leading LLM-based research agents and the base models, we demonstrate that Deep Research products can outperform base models in content overage and factual grounding, but still face challenges in hallucination, over-citation, etc. We hope that ReportBench will serve as a valuable tool for the research community to monitor, compare, and further improve the reliability of AI systems designed for academic survey tasks.

\bibliography{iclr2025_conference}
\bibliographystyle{iclr2025_conference}

\newpage

\appendix
\section{Appendix}

\subsection{Limitations}

ReportBench constructs 100 research tasks closely aligned with real-world scientific inquiry by reverse prompt engineering expert-written survey papers. It evaluates generated reports comprehensively along two axes: content quality and statement factuality. Despite its strengths, several limitations remain:

\textbf{Data Distribution.}
The benchmark primarily draws from peer-reviewed survey papers on arXiv, most of which are concentrated in STEM fields. This domain skew may limit the applicability of evaluations to other research areas. Future iterations will incorporate a broader set of source domains to improve coverage and generalization.

\textbf{Copyright Constraints.}
To mitigate legal risk, we only include papers under permissive licenses (CC BY 4.0, CC BY-SA 4.0, CC0 1.0, and the arXiv.org Non-exclusive license to distribute). The dataset is released under CC0 1.0 and contains only essential metadata (e.g., title, abstract, and references). Further narrowing the license scope would compromise domain balance. Authors who wish to opt out, please contact us for removal.

\subsection{Prompts in Evaluation}

\subsubsection{Cited Statement Extraction} 

\begin{tcolorbox}
[notitle,boxrule=0pt,colback=MyLightBlue!10,colframe=MyLightBlue!10]
\texttt{You are given a research report delimited by triple backticks.\\
Identify every statement that cites an external source (e.g. has a URL, DOI, or explicit citation marker) and pair it with the corresponding URL.\\
Return a JSON list where each item has two keys:\\
-  "statement": the single‑sentence claim, stripped of leading/trailing whitespace\\
-  "url": the canonical URL that supports that claim\\
If a citation contains multiple URLs, duplicate the statement for each URL.\\
ONLY return valid JSON.
Report:
```\{report\}```
}
\end{tcolorbox}

\subsubsection{Non-cited Statement Extraction} 

\begin{tcolorbox}
[notitle,boxrule=0pt,colback=MyLightBlue!10,colframe=MyLightBlue!10]
\texttt{You are given a research report delimited by triple backticks.\\
You are also given a list of statements that already have citations.\\
Your task is to identify factual claims or statements that:\\
1. Make specific assertions about facts, data, or events\\
2. Are NOT already included in the cited statements list\\
3. Could potentially be verified through external sources\\
4. Are NOT common knowledge or widely accepted facts\\
Exclude:\\
- Opinions, analysis, or subjective interpretations\\
- Statements that are already cited\\
- Common knowledge or universally accepted facts\\
- Vague or general statements\\
Return a JSON list where each item has one key:\\
- "statement": the factual claim that lacks citation support\\
ONLY return valid JSON.\\
Report:\\
```\{report\}```\\
Already cited statements:\\
\{cited\_statements\}\\
}
\end{tcolorbox}

\subsubsection{Supporting Source Extraction}

\begin{tcolorbox}
[notitle,boxrule=0pt,colback=MyLightBlue!10,colframe=MyLightBlue!10]
\texttt{You are provided with\\
Statement: \{statement\}\\
\\
Source Document:\\
\{source\_text\}\\
\\
Return any relevant content from the source document that supports the statement. This can be a sentence, paragraph, or even the entire text if necessary.\\
If no content supports it, return ``NOT\_FOUND''.\\
Return plain text only.\\
}
\end{tcolorbox}

\subsubsection{Semantic Consistency Verification} 

\begin{tcolorbox}
[notitle,boxrule=0pt,colback=MyLightBlue!10,colframe=MyLightBlue!10]
\texttt{You will decide whether a claim is correctly supported by a source sentence.\\
\\
Claim from report:\\
\{statement\}\\
\\
Source Sentence from original source:\\
\{source\_sentence\}\\
\\
Respond with JSON containing:\\
-  "reason": one short sentence explaining your decision\\
-  "match":  true or false          // true if the source sentence faithfully supports the claim\\
Return ONLY the JSON.
}
\end{tcolorbox}

\subsubsection{Web-based Statement Verification} 

\begin{tcolorbox}
[notitle,boxrule=0pt,colback=MyLightBlue!10,colframe=MyLightBlue!10]
\texttt{You are tasked with verifying the accuracy of a factual statement using web search capabilities.\\
\\
Statement to verify:\\
\{statement\}\\
\\
Please:\\
1. Use web search to find reliable, authoritative sources about this statement\\
2. Analyze the information you find from multiple sources\\
3. Determine if the statement is factually correct or incorrect based on your research\\
\\
Respond with JSON containing:\\
- "reason": a detailed explanation of your verification process and findings (2-3 sentences)\\
- "decision": true if the statement is correct, false if it is incorrect\\
\\
Only return the JSON response.
}
\end{tcolorbox}

\newpage

\subsubsection{Reference Title Extraction} 

\begin{tcolorbox}
[notitle,boxrule=0pt,colback=MyLightBlue!10,colframe=MyLightBlue!10]
\texttt{Please analyze the following academic survey and extract all cited academic paper titles and author information.\\
\\
Survey content:\\
\{response\}  \\
\\
Please reply in JSON format, containing an array named `papers`, where each paper object includes the following fields:\\
- title: the title of the paper\\
- authors: a list of authors\\
- is\_academic\_paper: true $($indicating this is an academic paper$)$\\
\\
Example format:\\
\{\\
\hspace*{1em}"papers": [\\
\hspace*{2em}\{\\
\hspace*{3em}"title": "Deep Learning for Natural Language Processing",\\
\hspace*{3em}"authors": ["John Smith", "Jane Doe"],\\
\hspace*{3em}"is\_academic\_paper": true\\
\hspace*{2em}\},\\
\hspace*{2em}...
\hspace*{1em}]\\
\}\\
\\
Note: Only extract explicitly mentioned academic papers. Do not include books, websites, or other types of references.\\
}
\end{tcolorbox}

\subsection{Prompt in Data Construction}\label{sec:app:promp_exmp}
\begin{tcolorbox}
[notitle,boxrule=0pt,colback=MyLightBlue!10,colframe=MyLightBlue!10]
\texttt{\fcolorbox{white}{HeadingColor!10}{\textbf{Sentence-level prompt}} \\
Please help me research the academic advancements in different radar data representation methods in the field of autonomous driving, and ensure only papers published before April 2025 are referenced.\\
\\
You also need to follow the following rules:\\
- Do not refer to the survey titled ``Exploring Radar Data Representations in Autonomous Driving: A Comprehensive Review''.\\
- Responses are given in the form of an English language survey with citations where appropriate.\\
\fcolorbox{white}{HeadingColor!10}{\textbf{Paragraph-level prompt}}\\
I am conducting a literature review on 3D LiDAR localization technology for autonomous vehicles. I hope you can summarize and analyze the major research directions and methods in this field, particularly methods based on 3D point cloud registration, methods based on 3D features, and emerging methods based on deep learning. Please ensure that all the referenced literature is published before November 2020.\\
\\
You also need to follow the following rules:\\
- Do not refer to the survey titled ``A Survey on 3D LiDAR Localization for Autonomous Vehicles''.\\
- Responses are given in the form of an English language survey with citations where appropriate.\\
\fcolorbox{white}{HeadingColor!10}{\textbf{Detail-rich prompt}}\\
I need a detailed academic research report on using Graph Neural Networks (GNN) for text classification. The report should systematically review advancements in this field, with a focus on the following aspects:\\
1. **Core Methodology**: Provide a detailed explanation and comparison of two main approaches: corpus-level GNNs and document-level GNNs. For each method, thoroughly analyze graph construction strategies (e.g., defining nodes and edges using PMI, TF-IDF, etc.), representation methods for nodes and edges, and graph learning algorithms (e.g., GCN, GAT, etc.).\\
2. **Key Model Analysis**: List and analyze representative models, such as TextGCN, SGC, BertGCN (corpus-level), and Text-Level-GNN, TextING (document-level).\\
3. **Evaluation and Challenges**: Summarize commonly used benchmark datasets in this field (e.g., 20NG, R8, MR) and evaluation metrics (e.g., Accuracy, F1-score), and discuss major challenges faced by current research, such as scalability, computational costs, and integration with pre-trained language models.\\
**Restrictions**:\\
- Only refer to and cite papers published **before July 2024**.\\
- Focus on English literature published in top conferences/journals in natural language processing and artificial intelligence (e.g., ACL, EMNLP, NAACL, AAAI, WWW, ICLR).\\
\\
You also need to follow the following rules:\\
- Do not refer to the survey titled ``Graph Neural Networks for Text Classification: A Survey''.\\
- Responses are given in the form of an English language survey with citations where appropriate.\\}
\end{tcolorbox}

\end{document}